\documentclass[sigconf]{acmart}
\usepackage{algorithm}
\usepackage{algpseudocode}
\algrenewcommand\textproc{\text}
\usepackage{makecell}
\usepackage{booktabs}
\usepackage{color}
\usepackage{multirow}
\usepackage{enumitem}
\usepackage{balance}
\usepackage{makecell}
\usepackage{threeparttable}
\usepackage{amsmath,amsfonts,mathtools}
\usepackage{mathrsfs}
\usepackage{float}
\usepackage{graphicx}
\usepackage{subfigure}
\usepackage{xcolor}
\usepackage{enumitem}

\AtBeginDocument{%
  \providecommand\BibTeX{{%
    \normalfont B\kern-0.5em{\scshape i\kern-0.25em b}\kern-0.8em\TeX}}}

\renewcommand\footnotetextcopyrightpermission[1]{} 
\acmConference[In submission to CIKM '23]{}{}

\begin{document}

\title{Uncertainty Quantification via Spatial-Temporal Tweedie Model for  Zero-inflated and Long-tail Travel Demand Prediction}

\author{Xinke Jiang}\authornote{Both authors contributed equally to this research.}
\affiliation{
\institution{Peking University}
\city{Beijing}
\country{China}
}
\email{thinkerjiang@foxmail.com}

\author{Dingyi Zhuang}
\authornotemark[1]
\affiliation{
\institution{Massachusetts Institute of Technology}
\city{Cambridge}
\state{Massachusetts}
\country{USA}
}
\email{dingyi@mit.edu}

\author{Xianghui Zhang}
\affiliation{%
  \institution{SpaceTimeLab\,University College London}
  \city{London}
  \country{United Kingdom}}
\email{xianghui.zhang.20@ucl.ac.uk}

\author{Hao Chen}
\affiliation{%
  \institution{School of Emergency Management Science and Engineering, University of Chinese Academy of Sciences}
  \city{Beijing}
  \country{China}}
\email{chenhao915@mails.ucas.ac.cn}

\author{Jiayuan Luo}
\affiliation{%
  \institution{
  Chengdu University of Technology}
  \city{Chengdu}
  \country{China}}
\email{joyingluo@foxmail.com}

\author{Xiaowei Gao}\authornote{Xiaowei Gao is the corresponding author.}
\affiliation{%
\institution{SpaceTimeLab\,University College London}
  \city{London}
  \country{United Kingdom}}
\email{xiaowei.gao.20@ucl.ac.uk}

\renewcommand{\shortauthors}{Jiang et al.}

\begin{abstract}
Understanding Origin-Destination (O-D) travel demand is crucial for transportation management. However, traditional spatial-temporal deep learning models grapple with addressing the sparse and long-tail characteristics in high-resolution O-D matrices and quantifying prediction uncertainty. This dilemma arises from the numerous zeros and over-dispersed demand patterns within these matrices, which challenge the Gaussian assumption inherent to deterministic deep learning models. To address these challenges, we propose a novel approach: the \underline{S}patial-\underline{T}emporal \underline{T}wee\underline{d}ie Graph Neural Network (STTD). The STTD introduces the Tweedie distribution as a compelling alternative to the traditional 'zero-inflated' model and leverages spatial and temporal embeddings to parameterize travel demand distributions. Our evaluations using real-world datasets highlight STTD's superiority in providing accurate predictions and precise confidence intervals, particularly in high-resolution scenarios. Anonymous GitHub code is available online\footnote{\url{https://github.com/STTDAnonymous/STTD}}.
\end{abstract}



\maketitle

\section{Introduction}


Efficient urban transportation hinges upon a balanced travel supply and demand, a balance greatly aided by accurate O-D travel demand forecasting \cite{ke2021predicting,xiong2020dynamic}. This precision in prediction allows for dynamic resource allocation, reducing wait times, and boosting service provider profitability \cite{zhang2019graph, jiang2022deep,liang2023region}. Thus, improving the model accuracy is the main focus of the travel demand prediction domain.


However, the task is not without its challenges, owing to the intricate spatial-temporal interdependencies and fluctuating nature of travel demand. While regions with dense demand, like airports and hospitals, generally present data that adheres to a Gaussian distribution—a core assumption of numerous prediction models—\cite{li2019estimating, kar2022essential}, the opposite is true for areas with sparse and discrete O-D demand, such as educational institutions and government premises \cite{niu2021delineating, chen2023sensing}. Such deviations from the Gaussian assumption further complicate forecasting. Moreover, the problem of data sparsity is exacerbated when accounting for the disparity in urban demand across different regions at high spatial-temporal resolutions, like at 5min intervals \cite{zhuang2022uncertainty}. A plethora of zero values, signifying the absence of trips, along with a long-tail distribution at higher demand levels, result in the skewness, discrepancy, and large variance in the data distribution \cite{wang2021low,wen2018estimation,zhang2020novel}. Hence, accurately interpreting zeros, capturing long-tail distributions, and understanding non-negative discrete values become paramount for robust demand forecasting.

Classic deep learning methods, such as Convolutional Neural Networks (CNNs) and Long Short-Term Memory networks (LSTMs), have tackled O-D matrix prediction by exploiting spatial and temporal dependencies \cite{Zhang2017DeepPrediction, liu2019contextualized, yao2018deep, wu2020comprehensive, SPGCL}. Recent advancements have introduced Graph Neural Networks (GNNs), which leverage the graph-like structure of O-D matrices to uncover non-Euclidean correlations \cite{Geng2019SpatiotemporalForecasting, xiong2020dynamic}. Despite their respective merits, all these models mainly treat O-D matrix entries as continuous variables, with a primary focus on coarse temporal resolutions. They usually simplify variance structures by assuming homoskedasticity (constant variance) and predominantly output expected average travel demand values. These approaches could overlook critical data features and may fail to sufficiently account for potential deviations and real-world uncertainties \cite{Rasouli2012UncertaintyAgenda, zhuang2022uncertainty}. Recent research \cite{wang2023uncertainty, zhuang2022uncertainty} suggests that integrating zero-inflation statistic models with deterministic deep learning frameworks might hold promise for modelling sporadic travel demand and quantifying uncertainty, even these approaches still fall short in adequately accounting for the long-tail distribution in sparse O-D demand.

In this paper, we propose the \underline{S}patial-\underline{T}emporal \underline{T}wee\underline{d}ie {G}raph {N}eural {N}etwork (STTD)—a comprehensive solution designed for joint numeric prediction and uncertainty quantification. Our main contributions can be summarized as follows:
\begin{itemize} [leftmargin=*]
    \item  We integrate the Tweedie distribution to model demand, replacing the traditional two-part zero-inflated model, thereby effectively capturing the zero-inflation and long-tail non-zero characteristics of O-D travel data.
    \item The proposed combination is adept at quantifying the spatial-temporal uncertainty inherent in sparse travel demand data.
    \item We validate the superiority of the STTD through experiments on two real-world travel demand datasets, tested across various spatial-temporal resolutions and performance metrics.
\end{itemize}

The paper is organized as follows. Section \ref{sec:method} defines the research question and develops the model. Section \ref{sec:experiment} introduces the dataset used for the case study, the evaluation metrics, and the experimental results. Section \ref{sec:conclusion} concludes the paper and discusses future research.

\begin{figure*}[t]
\vspace{-0.48cm}
    \centering  \includegraphics[width=0.90\linewidth]{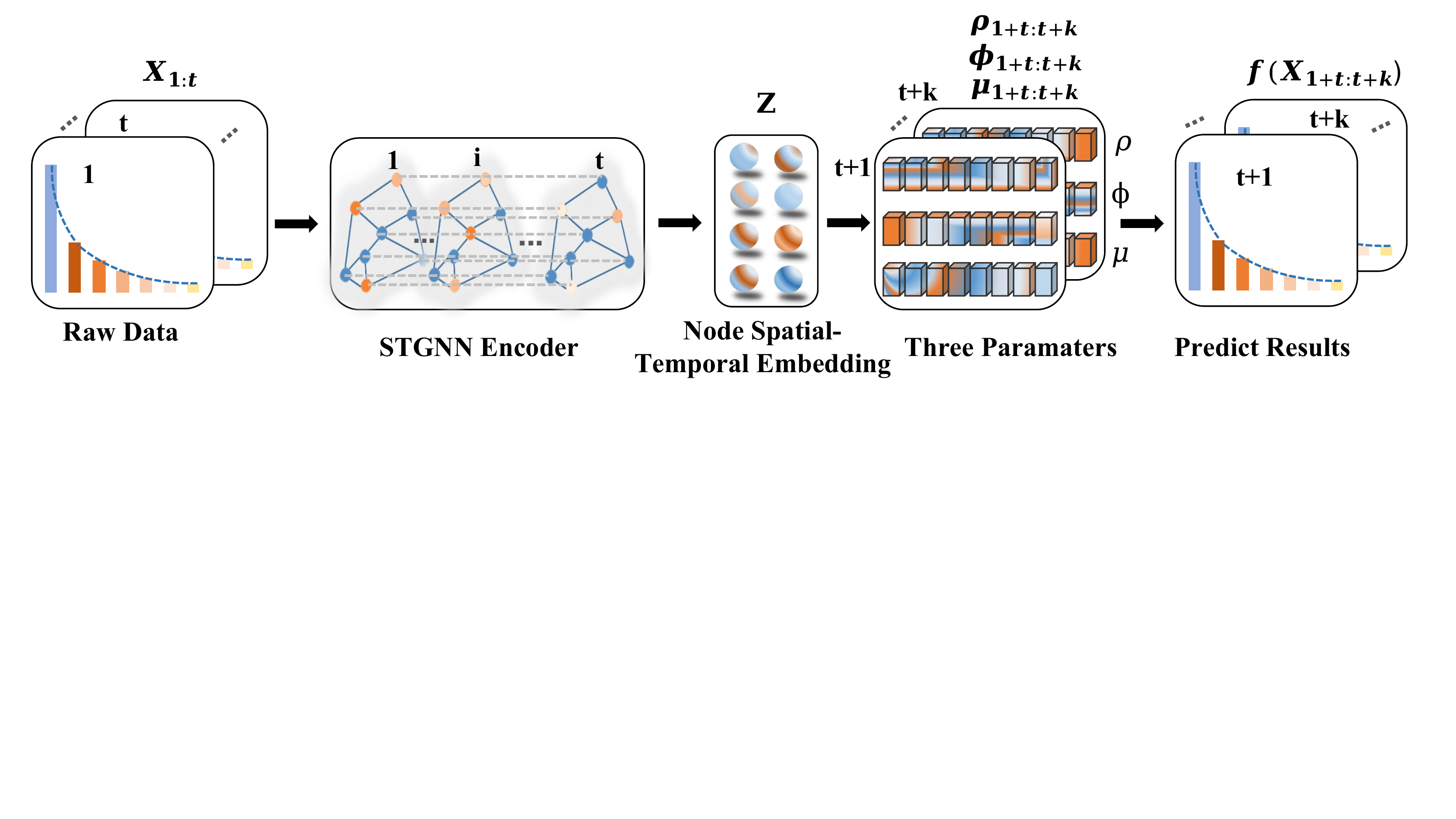}
    \vspace{-5.1cm}
\caption{Framework of STTD model.}
    \label{fig:model}
\end{figure*}

\section{Methodology}
\label{sec:method}
\subsection{Problem Description}
The primary objective of our model is to fit parameters that capture the future travel demand distributions for each O-D pair over a span of $k$ future time windows. The model accomplishes this by leveraging data from $m$ origins, $u$ destinations, and the corresponding travel demand within periods of length $T$ minutes, rendering the task essentially a sequence-to-sequence prediction. Unlike previous work that defined the locations of origins or destinations as vertices, we adopt a more effective approach and directly construct the O-D flow graph $\mathcal{G} = (V,E,A)$. Within this graph, $|V|=m\times u$ signifies the set of O-D pairs, $E$ designates the set of edges, and $A\in \mathbb{R}^{|V|\times|V|}$ represents the adjacency matrix that outlines the relationships among O-D pairs \cite{ke2021predicting,zhuang2022uncertainty}. Our approach provides a more nuanced understanding of the spatial-temporal intricate interrelationships present in urban travel demand.

We let $x_{it}$ represent the trips that occur at the $i^{th}$ O-D pair in the $t^{th}$ time window, where $i\in V$ and $x_{it}\in \mathbb{N}$. Our approach primarily considers the individual instances of travel demand at different intervals, all of which collectively model the term $x_{it}$. Subsequently, $X_{t}\in \mathbb{N}^{|V|\times T}$ designates the demand for all O-D pairs in the $t^{th}$ time window, with $x_{it}$ as its entry. The objective is to utilize historical records $X_{1:t}$ as the inputs for training data, aided by the graph structure $A$, to predict the probabilistic density function $f(X_{t+1:t+k})$ of the distribution of $X_{t+1:t+k}$—that is, the travel demand distribution for the next $k$ time windows. This prediction allows us to analyze the expected values and confidence intervals of travel demands.


\subsection{Tweedie (TD) Distribution}

Denote $f_{TD}$ as the probability mass function of future O-D travel demand $x_{it}$ out model outputs. It follows Tweedie distribution, which is in the form of: $f_{\text{TD}}(x_{it}|\theta, \phi) \equiv a(x_{it},\phi) \exp \left[ \frac{x_{it}\theta-\kappa(\theta)}{\phi} \right]$. Here, $\theta \in \mathbb{R}$ represents the natural parameter, while $\phi \in \mathbb{R}^+$ is the dispersion parameter. The normalizing functions $a(\cdot)$ and $\kappa(\cdot)$ correspond to parameters $\phi$ and $\theta$, respectively \cite{Tweedie2,td3}. Functions' details will be provided later. In the context of the Tweedie distribution, the mean and variance of a random variable $x$ are given by the following expressions:
$
E(x) = \mu = \kappa'(\theta),   Var(x) = \phi \kappa''(\theta),
\label{eq: td mean, var}
$
where $\kappa'(\theta)$ and $\kappa''(\theta)$ denote the first and second derivatives of $\kappa(\theta)$, respectively. Here, $\mu\geq 0$ is the mean parameter. The Tweedie family incorporates many significant distributions based on different index parameter $\rho$. This includes the Normal ($\rho=0$), Poisson ($\rho=1$), Gamma ($\rho=2$), Inverse Gaussian ($\rho=3$), and Compound Poisson-Gamma distribution ($1<\rho<2$) \cite{td4, td5, td3}. The Compound Poisson-Gamma distribution is particularly useful due to its ability to parameterize zero-inflated and long-tail data. When $1<\rho<2$, the demand $x_{it}$ can be expressed as shown in Eq.\ref{eq: y_sum}:

\begin{equation}
    \label{eq: y_sum}
    x_{it} = \left\{ \begin{array}{cc}
        0 & \text{ if no trips},  \\
        \sum_{j=1}^{L_{it}} l_{it}^{(j)} = l_{it}^{(1)} + l_{it}^{(2)} + \cdots + l_{it}^{(L_{it})} & \text{else} .
    \end{array}\right..
\end{equation}
where $L_{it}$, the number of time slices within the time window, follows a Poisson distribution $Pois(\lambda)$ with mean $\lambda$. The number of trips, $l_{it}^{(j)}$, are independent gamma random variables denoted by $Gamma(\alpha, \gamma)$ with mean $\alpha \gamma$ and variance $\alpha \gamma ^2$. In this way, $x_{it}$ is formed by the aggregation of discreet count, and we introduce $L_{it}$, $l_{it}^{(j)}$, and $x_{it} = \sum_{j=1}^{L_{it}} l_{it}^{(j)}$ to align with the Tweedie distribution definition\footnote {For implementation purposes, $x_{it}$ remains the model input. We clarify the specifics of $x_{it}$ to align with the structure necessitated by the Tweedie distribution}. If no trips occur, then $x_{it}=0$, and the probability mass at zero for travel demand is $P(x_{it}=0)=\exp{(-\lambda)}$ \cite{Tweedie2}. Otherwise, $x_{it}$ is computed as the sum of $L_{it}$ independent Gamma random variables. We re-parameterize the Tweedie distribution where $\theta = \mu^{1-\rho}/(1-\rho)$, and $\kappa(\theta)=\mu^{2-\rho}/(2-\rho)$ as:
\begin{equation}
\begin{aligned}
f_{\text{TD}}(x_{it}|\theta, \phi) \equiv
f_{\text{TD}}(x_{it}|\mu, \phi, \rho)
=a(x_{it}, \phi, \rho)e^ { \bigl[{x_{it} \frac{\mu^{1-\rho}}{\phi(1-\rho)}-\frac{\mu^{2-\rho}}{\phi(2-\rho)}} \bigl]}, \nonumber
\label{eq: re_f}
\end{aligned}
\end{equation}
with the normalizing function $a(x_{it}, \phi, \rho)$ defined as: $a(x_{it}, \phi, \rho) = \left\{ \begin{array}{cc}
        1 & \text{if } x_{it}=0,  \\
        \frac{1}{x_{it}} \sum_{j=1}^{\infty} 
        \frac{x_{it}^{-j \alpha}(\rho-1)^{\alpha j} }{\phi^{j(1-\alpha)} (2-\rho)^j j! \Gamma(-j\alpha)}
         & \text{if } x_{it}>0.
    \end{array}\right.
$. In this definition, $\mu, \phi$, and $\rho$ are the key parameters determining the probability and expected value of travel demands. The parameters in the Gamma and Poisson distributions, namely $\lambda, \alpha, \gamma$, can be computed using $\mu, \phi$, and $\rho$: $
\lambda = \frac{1}{\phi} \frac{\mu ^ {2-\rho}}{2-\rho}, \alpha = \frac{{2-\rho}}{\rho-1}, \gamma=\phi (\rho-1) \mu ^ {\rho-1}.
\label{eq: lambda, alpha, gamma}
$

The choice of the Tweedie distribution is driven by the characteristics of travel demand data. In practical terms, a specific time window may span different durations, such as 60 minutes, 15 minutes, or even as brief as 5 minutes. However, the distribution of trips during these periods can exhibit significant variations in both spatial and temporal dimensions. By incorporating more granular intervals into the model, it is possible to better capture the variability and heterogeneity of demand within the given time window. The Tweedie distribution effectively models zero-inflated and long-tail data distributions, making it particularly suited to this context.

\subsection{Learning Framework  and Loss Function}
We utilize Diffusion Graph Convolution Network (DGCN) and Temporal Convolutional Network (TCN) as Spatial-Temporal Graph Encoder $\mathcal{ST}$ \cite{zhuang2022uncertainty,wu2021inductive}. Thus, node spatial-temporal embedding $\mathcal{Z}$ can be denoted as: 
$
    \mathcal{Z} = \mathcal{ST}_{\Theta}(X_{1:t},A).
$
where $\mathcal{Z}_{i} \in \mathbb{R}^{F'}$ is the spatial-temporal embedding of the $i^{th}$ O-D pair. Thus, the three parameters $ \mu_{t+1:t+k}, \phi_{t+1:t+k}, \rho_{t+1:t+k}$ defining the Tweedie distribution can be computed as:
\begin{equation}
\centering
 \begin{aligned}
\mu_{t+1:t+k} &= \text{ReLU}(W_{\mu}\cdot \mathcal{Z} + b_{\mu}) \\
\phi_{t+1:t+k} &= \text{ReLU}(W_{\phi}\cdot \mathcal{Z} + b_{\phi}) + \epsilon \\
\rho_{t+1:t+k} &= 
\text{Sigmoid}(W_{\rho}\cdot \mathcal{Z} + b_{\rho}) + 1 + \epsilon
    \label{eq:four parameters at last layer}
 \end{aligned}
\end{equation}
where $ \mu \in [0, +\infty), \phi \in (0,+\infty), \rho \in (1,2)$. Learnable weight matrices $W_{\phi}, W_{\rho}, b_{\mu}, b_{\phi}, b_{\rho} \in \mathbb{R}^{F' \times k}$ and $\lim \epsilon \to 0$ is the minimum value. In order to fully predict travel demands, let $x^*$ be one of the predicted travel demand Tweedie distributions from $f_{TD}(X_{t+1:t+k})$ with parameters $\mu, \phi, \rho$ (notations $\mu, \phi, \rho$ are reused for clearer formula). The learning objective of the whole model can be represented as the maximum log-likelihood function: $\max \log f_{TD}(x^*|\mu, \phi, \rho)$ and directly use the negative likelihood as our loss function to better fit the distribution into the data. The log-likelihood of TD is composed of the  $x^*=0$ and $x^*>0$:
\begin{equation}
\begin{aligned}
    \label{eq: L}
    \mathcal{L}_{TD} = - \log f_{TD}(x^*>0|\mu, \phi, \rho) - \log f_{TD}(x^*=0|\mu, \phi, \rho) + \lambda{\Theta^2}.
    \nonumber
\end{aligned}
\end{equation}
where $\Theta$ is model paramaters and $\lambda$ is weight-parameter for L2 Normalization. Moreover, for $x^*>0$ :
\begin{equation}
\begin{aligned}
    \label{eq: max log n-z} 
    & \log f_{TD}(x^*>0|\mu, \phi, \rho)  
    = \frac{1}{\phi} \bigl (x^* \frac{\mu ^{1-\rho}}{1-\rho} - \frac{\mu ^{2-\rho}}{2-\rho}  \bigl) + \log a(x^*>0, \phi, \rho) \\
    & = \frac{1}{\phi} \bigl (x^* \frac{\mu ^{1-\rho}}{1-\rho} - \frac{\mu ^{2-\rho}}{2-\rho}  \bigl) - \log x^* + \log \sum_{j=1}^{\infty} 
        \frac{x^{*-j \alpha}(\rho-1)^{\alpha j} }{\phi^{j(1-\alpha)} (2-\rho)^j j! \Gamma(-j\alpha)} \\  
    & \geq \frac{1}{\phi} \bigl (x^* \frac{\mu ^{1-\rho}}{1-\rho} - \frac{\mu ^{2-\rho}}{2-\rho}  \bigl) - \log (j_{max} \sqrt{-\alpha} x^* ) + j_{max}(\alpha - 1) \nonumber
\end{aligned},
\end{equation}
where $j_{max}=\frac{x^{*2-\rho}}{(2-\rho)\phi}, \alpha = \frac{2-\rho}{1-\rho}<0$. As for $x^*=0$, 
$
    \label{eq: max log n-z2} 
    \log f_{TD}(x^*=0|\mu, \phi, \rho)
    =\frac{1}{\phi} \bigl ( - \frac{\mu ^{2-\rho}}{2-\rho}  \bigl)
$
, where $\mu, \phi, \rho$ are also selected and calculated according to the index of $x^*=0$ or $x^*>0$. We optimize the lower bound of $\mathcal{L}_{TD}$ during training to avoid calculation of summation formula. The whole framework is illustrated in Figure \ref{fig:model}.

\section{Numerical Experiments}
\label{sec:experiment}
\subsection{Experiment Setup}
\noindent  \textbf{{Datasets}}: Chicago Data Portal (CDP)\footnote{\url{https://data.cityofchicago.org/Transportation/Transportation-Network-Providers-Trips/m6dm-c72p}} and Smart Location Database (SLD)\footnote{\url{https://www1.nyc.gov/site/tlc/about/tlc-trip-record-data.page}} datasets. The CDP dataset includes trip records of ride-sharing companies in Chicago's 77 zones every 15 minutes, from 01/09/2019 to 30/12/2019. We randomly select $10\times 10$ grids of O-D pairs from CDP. The SLD dataset encapsulates For-Hire Vehicle trip records in 67 Manhattan administrative zones. We alter temporal resolution (5/15/ 60-minute intervals) and sample $10\times 10$ / $67\times 67$ O-D pairs to gauge our model's performance. Both datasets were used in previous work \cite{ke2021predicting,zhuang2022uncertainty}.

\noindent  \textbf{{Evaluation Metrics}}: (1) Point estimates: Mean Absolute Error (MAE), which measures the accuracy of the mean or median value of the predicted Tweedie distributions. (2) Distributional uncertainty: Mean Prediction Interval Width (MPIW) and Prediction Interval Coverage Probability (PICP) within the 10\%-90\% confidence interval. MPIW averages the width of the confidence interval, while PICP quantifies the percentage of actual data points within confidence intervals. Additionally, KL-Divergence is applied to evaluate the similarity between the predicted and real data distributions. (3) Discrete demand prediction: true-zero rate and F1-score. The true-zero rate measures the model's fidelity in reproducing data sparsity, and the F1-score gauges the accuracy of discrete predictions. 

In general, Lower MPIW and KL-Divergence values are favorable while larger true-zero rate, PICP, and F1-score values denote superior model performance.


\noindent  \textbf{{Baselines}}: {Historical Average (HA)}, Spatial-Temporal Graph Convolutional Networks (STGCN)\footnote{\url{https://github.com/FelixOpolka/STGCN-PyTorch}}\cite{Yu2018Spatio-TemporalForecasting}, Spatial-Temporal Graph Attention Networks (STGAT), the state-of-the-art probabilistic models under zero-inflated negative binomial assumptions (STZINB) and other methods such as negative binomial (STNB), truncated normal (STTN) \footnote{\url{https://github.com/ZhuangDingyi/STZINB}}. We also evaluate our methods with different index parameter $\rho$ including $\rho \in [1,2]$ (STTD), $\rho=0$ (STG), $\rho=1$ (STP), $\rho=2$ (STGM) and $\rho=3$ (STIG). Note that STTD and STZINB are three-parameter models while other probabilistic components are two-parameter models.

\noindent  \textbf{{Reproducibility}}: Parameters of baselines are optimized using the Adam Optimizer\cite{Adam} with $L_2$ regularization and a dropout rate of 0.2. The GNN in STTD and baselines are all two-layered with hidden unit equal 42. 
We also employ the early-stopping strategy with patience equals 10 to avoid over-fitting. 
We split the data into 60\% for training, 10\% for validation, and the remaining 30\% for testing. All graph adjacency matrices are built following \cite{zhuang2022uncertainty}. 

\subsection{Model Comparison}
\label{sec:model_comparison}

\begin{table*}[!t]
\footnotesize
\caption{Model comparison under different metrics. $X/Y$ values correspond to the mean/median values of the distribution.}
\label{tab:comparison}
\setlength\tabcolsep{3.5pt}
\centering
\begin{tabular}{cc|c c c c c| c c c | c c c}
\toprule
Dataset & Metrics & \makecell{STTD \\ ($1<\rho<2$)}
  & \makecell{STG  \\ ($\rho=0$)} & \makecell{STP\\ ($\rho=1$)} & \makecell{STGM \\($\rho=2$)} & \makecell{STIG\\($\rho=3$)} & STZINB & STNB  & STTN & STGCN & STGAT & HA \\
\midrule
\multirow{5}{*}{CDPSAMP10}&MAE   &  \textbf{0.358}/\underline{0.363}  & 0.409/0.409 & 0.429/0.453 & 0.479/0.472&  0.376/0.384   & 0.368/{0.366} & 0.382/0.379 &  0.432/0.606  & 0.395 & 0.397 & 0.522\\
&MPIW & \textbf{0.059} & 2.407 & \underline{0.758} & 1.154 & 1.032 &  {1.018} & 1.020  & 2.089 & /\ & / &/\\\
&PICP  & \textbf{0.976}  & 0.874 & 0.948 & 0.792  & 0.865 & \underline{0.958} &  0.957 & 0.065 & /\ &  / & / \ \\
&KL-Divergence & 0.184/0.228& 0.435/0.435 &0.138/0.157 &0.120/0.128&\underline{0.113}/\textbf{0.106} & {0.291}/0.424 & 0.342/0.478 &  1.058/0.928  & 0.897 & 1.169 & 1.377\\
&True-zero rate &\textbf{0.803}/\underline{0.799} & 0.790/0.790 &0.503/0.552 &0.791/0.797&0.797/0.797 &0.796/0.788 & 0.796/0.788 &  0.758/0.764  &  {0.800}& 0.787 & 0.759 \\
&F1-Score  &\textbf{0.862}/0.844 & 0.818/0.818  &0.619/0.642 &0.832/0.843&  0.822/0.836& \underline{0.848}/0.846 &  \underline{0.848}/0.841  & 0.842/0.846  & 0.840& 0.846 &  0.809 \\

\midrule

\multirow{5}{*}{SLDSAMP10}&MAE &0.648/0.670 & \underline{0.627}/0.630 & 0.649/0.654& 0.658/0.660 & 0.638/0.640&0.663/0.666 & \underline{0.627}/ \textbf{{0.616}} & 0.695/0.665   & 0.630 & 0.678 & 0.697 \\
&MPIW    &\textbf{0.965} &  2.604 & \underline{1.063} & 1.523 & 1.466 &  {1.310} & 3.628 &  1.931  & /\ & / & / \ \\
&PICP    & 0.921 & 0.811 & 0.907 & 0.481 & 0.502 & \underline{0.942} & \textbf{0.943} & 0.219 & /\ & / & / \ \\
&KL-Divergence &\textbf{0.102}/0.113  & 1.022/1.022 & \underline{0.111}/0.122 & 0.130/0.130& 0.123/0.113&0.518/ {0.507} & 0.980/1.662 &  3.578/3.052   & 0.768 & 0.754 & 0.978 \\
&True-zero rate &\textbf{0.545}/0.529 &0.461/0.461 &0.504/0.508&0.522/\underline{0.531}&0.507/0.503 &0.499/ {0.502} & 0.465/0.418 &   0.308/0.336   & 0.478 & 0.508 & 0.364 \\
&F1-Score  &0.611/0.605 & 0.555/0.555 & 0.568/0.568& \underline{0.659}/\textbf{0.661}& 0.657/0.653& {0.567}/0.566 & 0.556/0.552 &   0.477/0.500   & 0.563 & 0.498 & 0.456 \\

\midrule
\multirow{5}{*}{SLD\_5min}&MAE & \textbf{0.139}/0.147&0.155/0.155 & 0.146/0.149 & 0.145/\textbf{0.139} & \underline{0.142}/0.143 & 0.149/0.150 &  0.147/ {0.144}   & 0.155/0.155  & 0.159 & 0.162 & 0.149 \\
&MPIW  & \underline{0.031} & 0.922 & \textbf{0.016} & 0.911&   1.224&  {0.094} & 1.249    & 0.741  & /\ &  / & / \ \\
&PICP  & \underline{0.973} & 0.895 & \textbf{0.981} & 0.924  & 0.888 & 0.968 &  0.969 & 0.037 & /\ &  / & / \ \\
&KL-Divergence &\textbf{0.001}/\textbf{0.001} &  {\textbf{0.001}/\textbf{0.001}} & 0.003/0.004& 0.004/0.003 & 0.003/\underline{0.002} & 0.015/0.014  & 0.042/0.145 &   \textbf{0.001}/\textbf{0.001}   & 0.056  & 0.053 &  0.060 \\
&True-zero rate &\textbf{0.884}/\underline{0.883} & 0.877/0.877 & 0.880/0.879 & 0.878/0.872 & 0.866/0.871&  {0.879/0.879} & 0.875/0.866 &   0.877/0.877  & 0.874 &  0.851 &  0.874 \\
&F1-Score & \textbf{0.905}/\underline{0.894}&  0.879/0.879 & 0.884/0.839 & 0.884/0.882 & 0.884/0.884 &  {0.882/0.882} & 0.880/0.878 &   0.879/0.879   & 0.879  & 0.868 & 0.876 \\

\midrule

\multirow{5}{*}{SLD\_15min}&MAE & \textbf{0.337}/0.354&  0.356/0.356 & 0.357/0.360&0.362/0.361 &0.360/0.360 & 0.370/0.372 & 0.351/ \underline{0.342} &   0.365/0.356 & 0.373 & 0.394  &     0.418 \\
&MPIW   & \textbf{0.141}  & 1.353  & 1.153 & 0.623 & 0.648 & \underline{0.603} & 2.283 &   1.215   & /\ & /\ & / \\
&PICP  & 0.938 & 0.855 & \textbf{0.962} & 0.724  & 0.715  & 0.956 & \underline{0.959}  &  0.120 & /\ &  / & / \ \\
&KL-Divergence   & \textbf{0.058}/\underline{0.061}  & 0.353/0.353 & \underline{0.061}/0.064& 0.063/0.067&0.062/0.071&0.167/ {0.156} & 0.357/0.704 &    1.445/1.211  & 0.395  &  0.434  & 0.445 \\
&True-zero rate & 0.727/\textbf{0.729}& 0.709/0.709 & \underline{0.728}/0.721&0.722/0.712&0.724/0.701 &0.725/ {0.727} & 0.710/0.684 &    0.632/0.648  & 0.708 & 0.704 & 0.703  \\
&F1-Score    & \textbf{0.796}/0.778   &0.750/0.750&0.776/0.722 &0.774/0.775  &0.772/\underline{0.786} & {0.751}/0.750 & 0.746/0.745 &    0.716/0.726  & 0.750 & 0.753 & 0.744 \\

\midrule

\multirow{5}{*}{SLD\_60min}&MAE & \textbf{0.915}/\underline{0.928} &1.199/1.199 & 0.942/0.956 & 0.937/0.971& 0.952/0.933& 1.040/1.067 & 0.958/ {0.947} &   1.275/1.254   & 0.997 & 0.987 & 1.014 \\
&MPIW    & 2.241 & 2.282 & 2.576 & \textbf{1.372} & \underline{1.431} & 3.277 & 5.753 &   {1.592}  & /\ & /\ & / \\
&PICP  & \textbf{0.964} & 0.557 & 0.872  & 0.524  & 0.533 & \underline{0.947} &  0.930 & 0.133 & /\ &  / & / \ \\
&KL-Divergence & \textbf{0.337}/0.395 & 2.176/2.176 &0.378/0.421 &0.363/0.342& 0.365/\underline{0.352}&0.982/1.270 &  {0.926}/0.963 &   4.120/3.734   & 1.114 & 2.053 &  2.421 \\
&True-zero rate & \textbf{0.505}/\underline{0.488}  & 0.390/0.390 & 0.474/0.429&0.453/0.449&0.354/0.379& 0.458/ {0.476} & 0.443/0.425 &   0.288/0.308   & 0.438 & 0.416 &  0.447 \\
&F1-Score   &  \textbf{0.617}/0.608 & 0.479/0.479 & 0.610/0.569 & \underline{0.613}/0.609& 0.604/0.615 &0.536/0.537 &  {0.538}/0.534 &   0.407/0.423  &  {0.538} & 0.488 &   0.490 \\
\bottomrule
\end{tabular}
\end{table*}

We carry out experiments on five distinct travel demand scenarios, presenting the prediction results in Table \ref{tab:comparison}. Here, the best and second-best scores are highlighted with bold and underlined values, respectively. As a general observation, STTD outperforms all baseline methods in the majority of cases when $\rho\in [1,2]$. Other models from the Tweedie family tend to surpass probabilistic methods while also exceeding other state-of-the-art deep learning techniques. For instance, STTD's numerical accuracy improvements reach up to 3.47\% on SLDSAMP10 in terms of Mean Absolute Error (MAE).
Moreover, STTD realizes improvements in Mean Predictive Interval Width (MPIW) of 327.66\%, 92.22\%, and 67.02\% on SLD\_15min, CDPSAMP10, and SLD\_5min scenarios respectively, when compared to STZINB and STNB. This suggests that STTD is capable of generating more precise predictions while maintaining a narrower confidence interval, a conclusion also supported by the Prediction Interval Coverage Probability (PICP) values. Typically, a model necessitating a larger confidence interval is indicative of a higher PICP, aiming to encapsulate all ground truth data points. However, STTD manages to sustain strong performance in both PICP and MPIW, implying the generated confidence intervals adeptly capture the underlying data distribution while remaining relatively narrow. In terms of capturing sparsity, STTD surpasses other methods on metrics such as KL-Divergence, true-zero rate, and F1-Score in most cases, signifying STTD's efficiency in capturing zero values. This attribute is vital for sparse travel demand prediction tasks where accurate and confident prediction results are paramount for effective decision-making. Three-parameter models like STTD and STZINB outshine two-parameter models and 'one-parameter' methods such as STGCN and STGAT. It's important to note that no single model dominates others across all resolution levels, as per Wolpert's "No Free Lunch" theorem \citep{wolpert1997no}.

\begin{figure}[htbp]
\vspace{-0.2cm}
  \centering
  \hspace{-1.0cm}
  \includegraphics[scale=0.29]{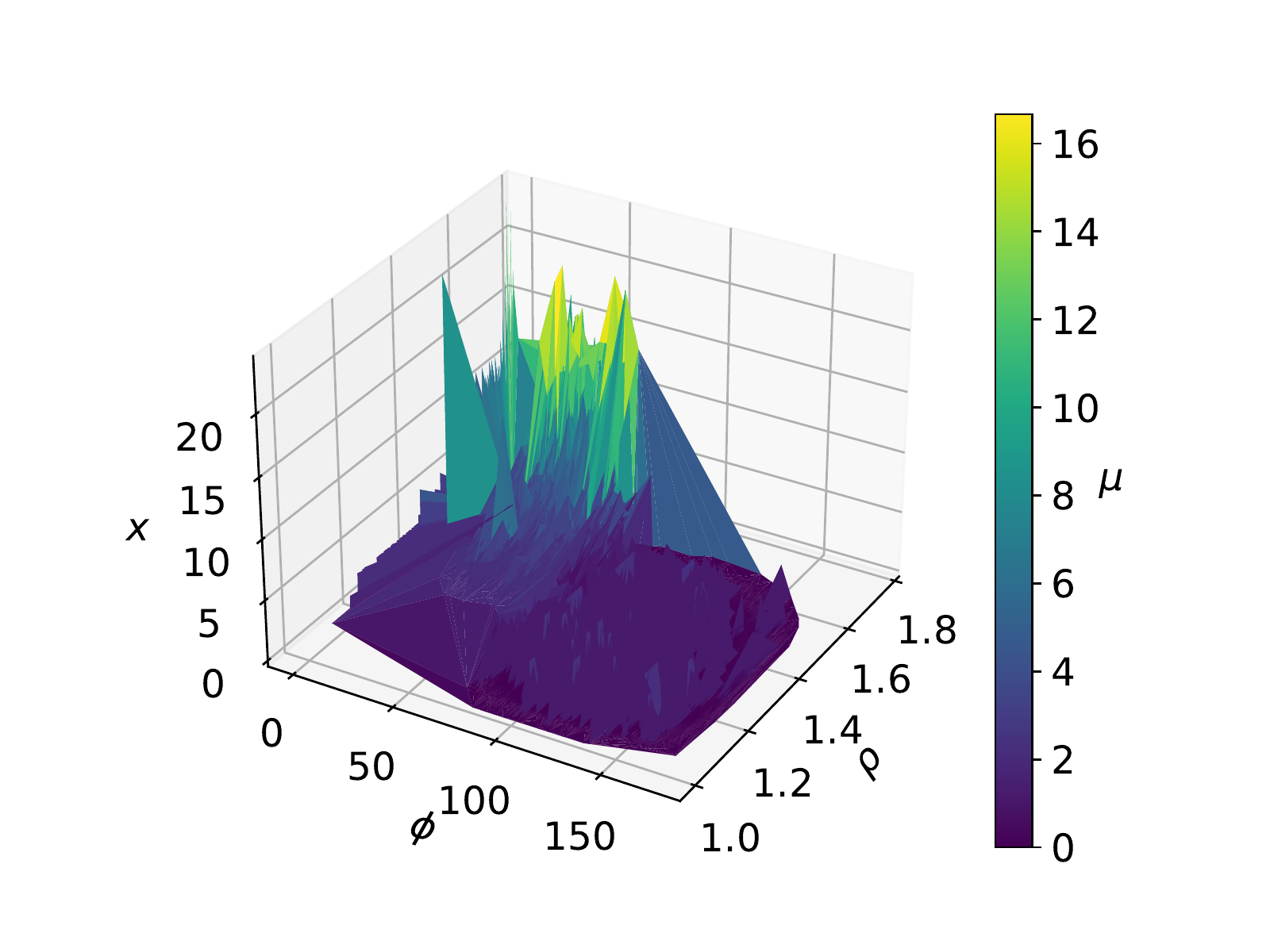}
  \hspace{-0.6cm}
  \includegraphics[scale=0.29]{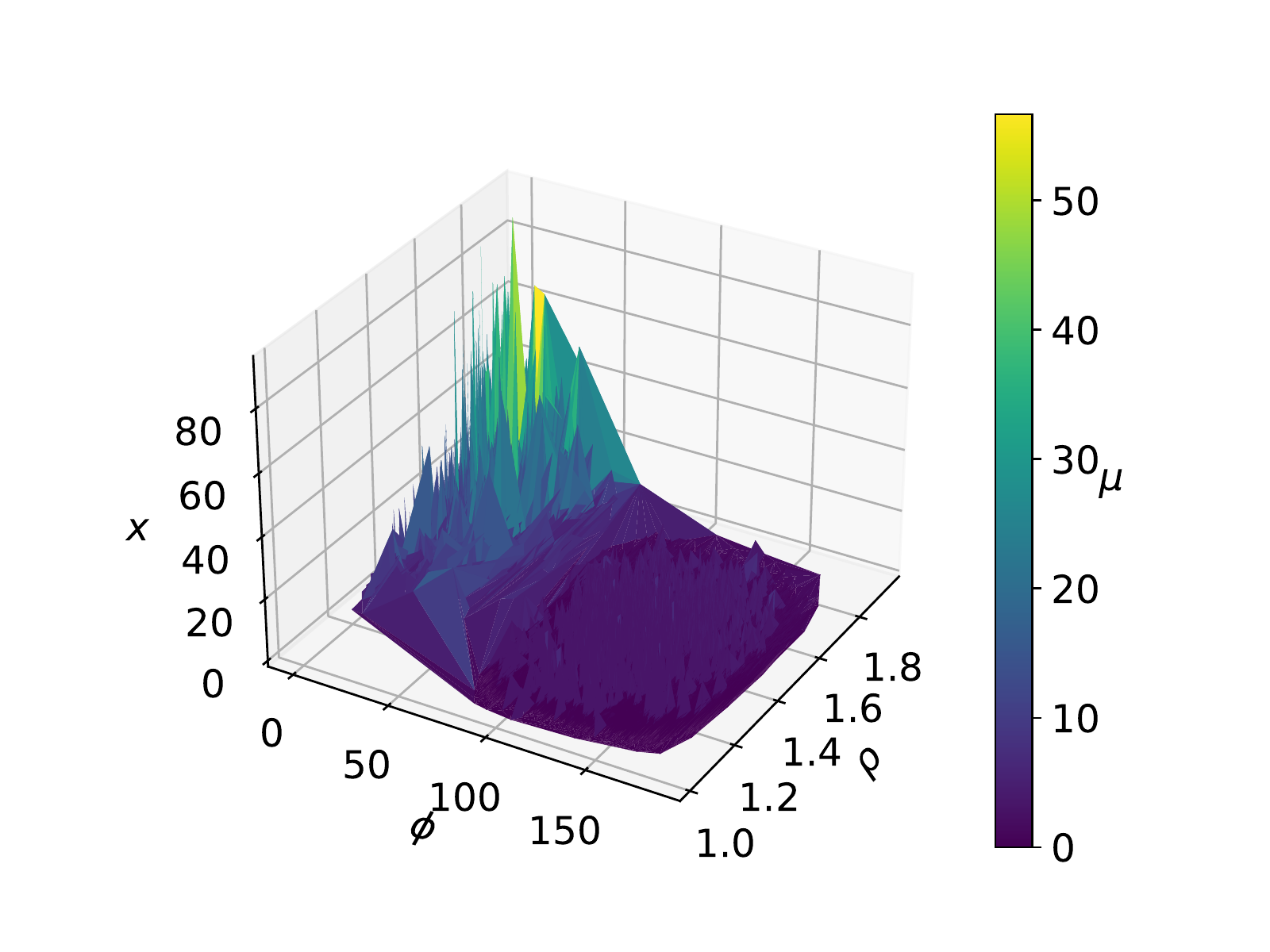}
  \hspace{-1.2cm}
  \caption{Surface plots for learned STTD parameters ($\mu,\phi,\rho$) on CDPSAMP10 (left) and SLDSAMP10 (right) test sets.
  }\label{fig:decouple_coff}
\end{figure}

\subsection{Parameters Visualization}
\label{sec:uq}

We visualize the learned parameters $\phi, \rho, \mu$ and real values $x$ by 3D surface plots in Figure \ref{fig:decouple_coff} on CDPSAMP10 and SLDSAMP10 test sets. As the plots provided, it is evident that for long-tailed data, the learned values of $\rho$ are greater (closer to 2), and a reason stands that the main part of loss function $x^*\frac{\mu ^ {1-\rho}}{1-\rho} (x^*>0)$ will be punished when predicted $x^*$ lies in long-tail while the predict approaches zero, making the loss function converts sharply. While for zero-valued data, the learned parameter $\phi$ is huge. Thus it shows that the model can capture zero-inflated and long-tail data effectively from uncertainty aspect. Furthermore, as the model parameter $\mu$ serves as the distribution mean, its color distribution matches the value of $x$, indicating that the model has a good performance in mean point estimation.


\section{Conclusion}
\label{sec:conclusion}
In this work, we introduced the Spatial-Temporal Tweedie Distribution (STTD), a novel neural probabilistic graph-based deep learning model aimed at effectively quantifying spatial-temporal uncertainty. STTD leverages the power of the Tweedie distribution, adept at handling zero-inflated and long-tail data. In conjunction with this, we utilize the capabilities of Spatial-Temporal Graph Neural Networks to effectively encode three key uncertainty parameters. This strategy allows STTD to capture the complex, intertwined spatial-temporal dependencies, and intrinsic uncertainties at each data point. We validated the performance of our model through extensive experiments across five representative scenarios, with a keen focus on point estimation and uncertainty measurement. Our results underscore the model's robustness and effectiveness, setting a new benchmark in the field.

Additionally, our approach brings to light the significance of rigorous uncertainty quantification within spatial-temporal deep learning frameworks, providing a robust platform for future research and development and also contributing to more efficient and reliable transport management systems.


    


\balance
\bibliographystyle{ACM-Reference-Format}
\bibliography{acmart}

\end{document}